\documentclass{article}

\PassOptionsToPackage{sort, numbers}{natbib}

\usepackage[final]{nips_2018}

\usepackage[utf8]{inputenc} 
\usepackage[T1]{fontenc}    
\usepackage{hyperref}       
\usepackage{url}            
\usepackage{booktabs}       
\usepackage{amsfonts}       
\usepackage{nicefrac}       
\usepackage{microtype}      

\usepackage{amsmath}
\usepackage{amssymb}
\usepackage{graphicx}
\usepackage{subfigure}
\usepackage{array,multirow}
\usepackage{float}
\usepackage{xspace}
\usepackage{bm}
\usepackage{makecell}

\newcommand{\bx}{{\mathbf{x}}}
\newcommand{\by}{{\mathbf{y}}}

\newcommand{\comment}[1]{}

\makeatletter
\DeclareRobustCommand\onedot{\futurelet\@let@token\@onedot}
\def\@onedot{\ifx\@let@token.\else.\null\fi\xspace}
\def\eg{\emph{e.g}\onedot} 
\def\ie{\emph{i.e}\onedot}

\makeatother

\title{Non-local RoI for Cross-Object Perception}

\author{
  Shou-Yao Roy Tseng \\
  National Tsing Hua University, Taiwan\\
  \And
   Hwann-Tzong Chen\thanks{{\bf Acknowledgement}. This work was supported in part by the MOST, Taiwan under Grants 107-2634-F-001-002 and 106-2221-E-007-080-MY3.} \\
  National Tsing Hua University, Taiwan\\
  \AND
  Shao-Heng Tai \\
  Umbo Computer Vision \\
  \And
  Tyng-Luh Liu$^*$ \\
  Academia Sinica, Taiwan \\
}

\begin{document}

\maketitle

\begin{abstract}
 We present a generic and flexible module that encodes region proposals by both their intrinsic features and the extrinsic correlations to the others. The proposed {\em non-local region of interest} (NL-RoI) can be seamlessly adapted into different generalized R-CNN architectures to better address various perception tasks. Observe that existing techniques from R-CNN treat RoIs independently and perform the prediction solely based on image features within each region proposal. However, the pairwise relationships between proposals could further provide useful information for detection and segmentation. NL-RoI is thus formulated to enrich each RoI representation with the information from all other RoIs, and yield a simple, low-cost, yet effective module for region-based convolutional networks. Our experimental results show that NL-RoI can improve the performance of Faster/Mask R-CNN for object detection and instance segmentation.  
\end{abstract}

\section{Introduction}
The current trend of deep network architectures for object detection can be categorized into  one-stage detectors and two-stage detectors. One-stage detectors perform the task of object detection in an end-to-end single-pass manner, \eg YOLO~\citep{Joseph2016YOLO, Joseph2017YOLOv2, Redmon2018YOLOv3} and SSD~\citep{Liu2016SSD, Fu2017DSSD}. On the other hand, two-stage detectors divide the task into two sub-problems that respectively focus on extracting object region proposals and classifying each of the candidate regions. Detectors such as Faster R-CNN~\cite{Ren2015Faster} and Light-Head R-CNN~\cite{Li2017lighthead} are both of this kind.

State-of-the-art object detection methods~\citep{HeZR2014SPP,Ross2015FastRCNN,Ren2015Faster,DaiLHS16,DaiQXLZHW17,Lin2017FPN,He2017MaskRCNN} in terms of precision mainly follow the {\em region based} paradigm, which is popularized by the seminal work R-CNN~\citep{Ross2014RCNN}. Given a sparse set of region proposals, object classification and bounding box regression are performed on each proposal individually.
Mask R-CNN~\citep{He2017MaskRCNN} extends Faster R-CNN by adding a branch for predicting segmentation masks on each Region of Interest (RoI) in parallel with the existing branch for classification and bounding box regression. This showcases the flexibility of two-stage detectors for multitasking over the one-stage counterparts. Different branches in Mask R-CNN share the same set of high-level features from a CNN backbone network, such as ResNet~\citep{He2016ResNet}. Each branch attends to a specific RoI via {\em RoIAlign}, a quantization-free layer that faithfully preserves spatial preciseness. Further, our {\em non-local RoI} (NL-RoI) mechanism can be incorporated into Mask R-CNN to achieve better performance.

The ability to capture long-range and non-local information is a key success factor of deeper CNNs. For vanilla Mask R-CNN, the only way to acquire non-local information for each RoI is to explore the high-level features extracted by the deep backbone network. However, the high-level features are shared among all RoIs of different spatial locations, semantic categories, and branches for different tasks. Such high-level features are assumed to be general rather than specific for individual RoIs so that they are applicable to all the above varieties. Therefore, it is difficult for the same set of features to also contain the RoI-specific information. Besides, RoI features are rectangularly extracted based on their corresponding bounding box proposed by the \textit{Region Proposal Network (RPN)}. It is very likely to have multiple instances in a single bounding box when the scene is crowded. Moreover, if the instances are of the same category, it is harder for the branch network to tell apart the boundary by only referring to the local feature within an RoI. Especially for non-rigid objects, such as persons, the target object will deform in shape, and the bounding box has a higher chance to include other objects or backgrounds interlacing in a more complicated way.

We introduce the idea of NL-RoI to address the aforementioned issues and argue that RoI-specific non-local information can be helpful in discriminating the target instance from the others. For example, due to object co-occurrence prior in the real world, it is more probable to see cars along with pedestrians instead of refrigerators in a street scene. Besides, mutual information between instances may also be useful. Consider a scene of group dancing: People are usually posing in similar ways, and hence we can more confidently predict the pose for a dancer under partial occlusion, by referring to other dancers' poses.

\section{Formulation}
Inspired by the non-local operation in \citep{Wang2018Nonlocal}, we define a generic non-local RoI operation for the use in conjunction with R-CNN based models \cite{Ross2014RCNN}:
\begin{equation}
\by_{i} = \frac{1}{\bm{C}(X)_i} \sum_{j=1}^{N}{ f(\mathbf{x}_{i}, \bx_{j}) g(\bx_{j})} \,,
\label{equation:nonlocal_roi}
\end{equation}
\noindent
where $i$ is the index of a target RoI whose non-local information is to be computed and $j$ enumerates all the $N$ RoIs, including the target one. The input feature blob is denoted as $X=[\bx_1,\cdots,\bx_N]$ and the output feature containing non-local information is denoted by $Y=[\by_1,\cdots,\by_N]$. A pairwise function $f$ computes a scalar that reflects the correlation between the $i$th target RoI and each of the RoIs ($\forall{j} \in \{1..N\}$). The unary function $g$ maps the input feature from the $j$th RoI to another representation, which gives the operation the capacity to convert the input feature to be more specialized for non-local information. Finally, the response is normalized by a factor $\bm{C}(X)_i
= \sum_{j=1}^N{f(\bx_{i}, \bx_{j})} $.

\begin{figure}
  \centering
  (a) \includegraphics[width=0.37\linewidth]{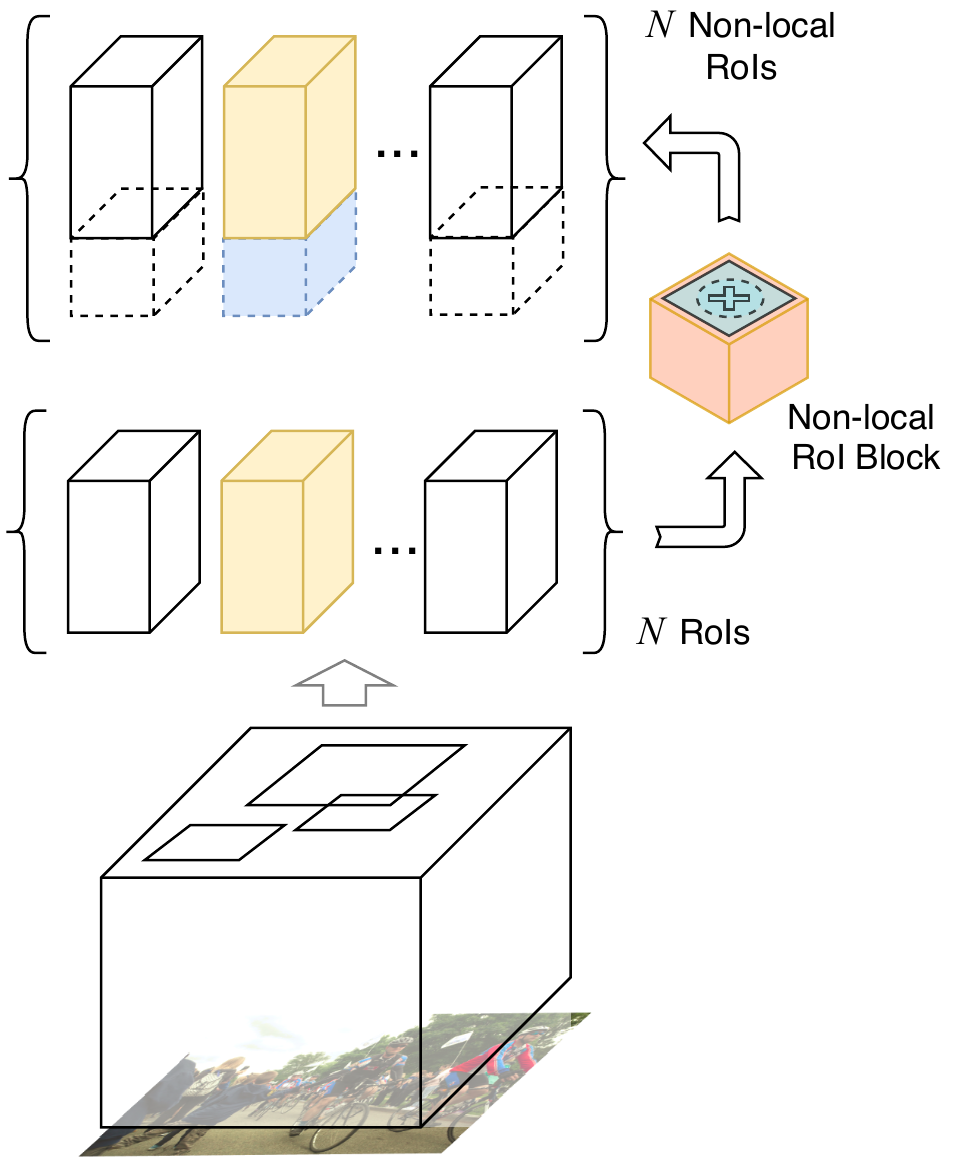} 
  (b) \includegraphics[width=0.5\linewidth]{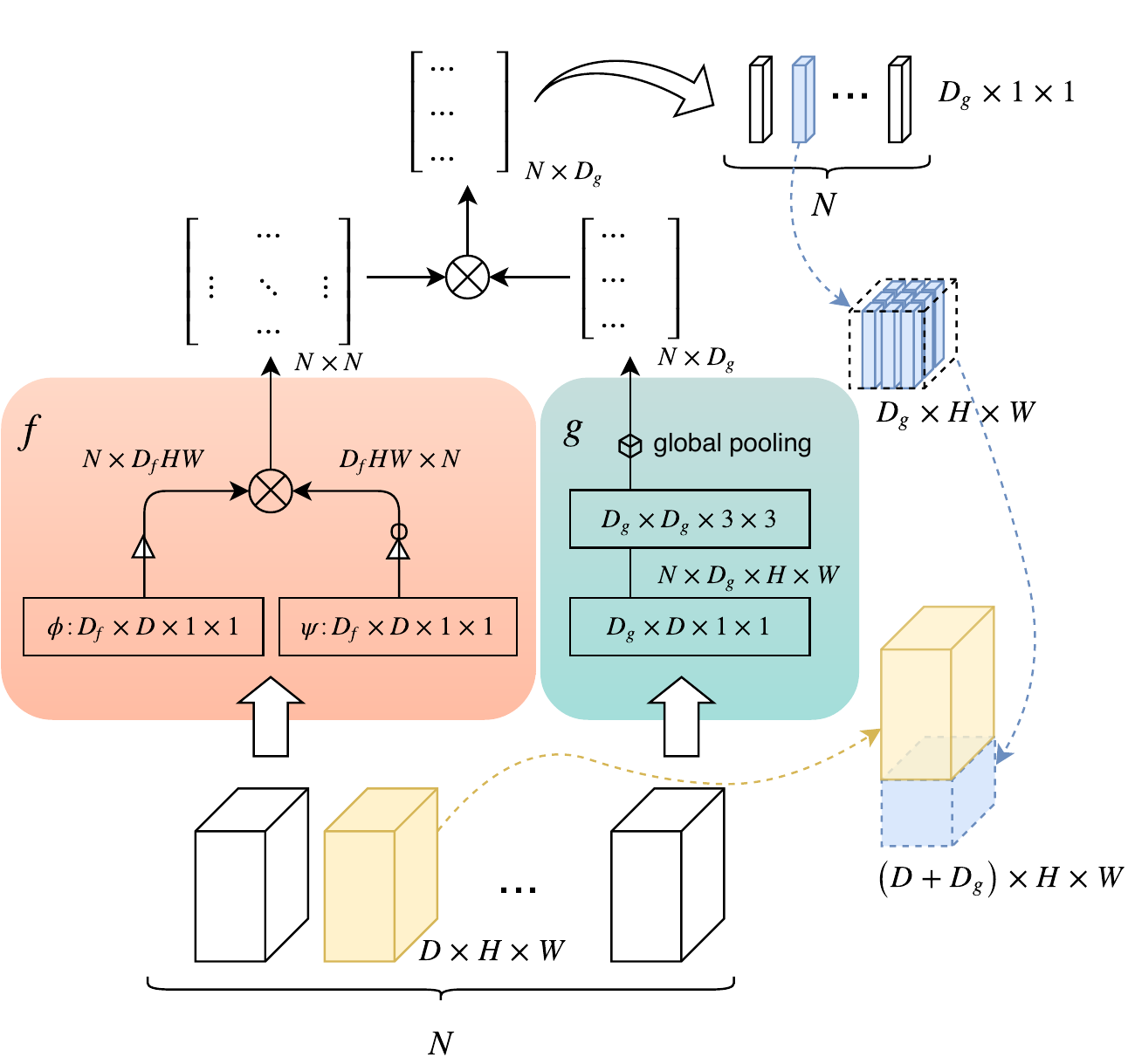} 
  \caption{(a) The yellow block represents the original high-level feature tensor extracted by the backbone network. The blue block represents the non-local feature calculated by the NL-RoI module. The original RoI feature tensor is concatenated with the non-local feature tensor.
  (b) The detailed operations of an NL-RoI module. Assume $N$ RoIs are proposed. The relation function $f$ computes scores on two flatten features obtained by two functions $\phi$ and $\psi$, which are 1-by-1 convolution layers. The embedding function $g$ consists of two convolution layers and a pooling layer. The first convolution layer is designed to lower down the feature channel dimension, and the second one uses 3-by-3 kernels to extract non-local specific features. The final non-local features obtained by relation scores and the embedded features produced by $g$ are one-dimensional vectors for each RoI. These one-dimensional features are tiled around $H,W$ spatial dimensions to form a three-dimensional tensor, and then are concatenated with original RoI feature tensors.
  }
  \label{fig:NLB}
\end{figure}

The non-local RoI property in Eq.~(\ref{equation:nonlocal_roi}) originates from the fact that all RoIs are associated with each other in the operation. For each RoI, the non-local RoI operation computes responses based on correlations between different RoIs. Theoretically, each RoI should gradually learn to characterize a meaningful instance during training. That is, Eq.~(\ref{equation:nonlocal_roi}) enables the attention mechanism between instances. Moreover, this kind of non-local operation supports a variable input number $N$ of RoIs. Fig.~\ref{fig:NLB}a shows an overview of NL-RoI.

\section{Implementation}
For simplicity, we adopt the {\em Embedded Gaussian} version of $f$:
$f(\bx_{i},\bx_{j}) = e^{\phi({\bx_{i}})^{T} \psi({\bx_{j}})}$.
Assume that we have $N$ RoIs and $D$ channels of input features, and the aligned RoI spatial size is $H \times  W$. Hence, the input feature blob $X$ has the shape of $(N, D, H, W)$. The two embedding functions $\phi$ and $\psi$ are both chosen to be a 1-by-1 2D convolution that reduces the channel dimension of the input blob. The purpose of $f$ is to calculate the correlations between $N$ RoIs, so the output of $f$ being applied to the whole input blob $X$ should be an $N$-by-$N$ matrix. The output blobs from $\phi$ and $\psi$ are reshaped to $(N, D_f \times H \times W)$. Afterward, a matrix multiplication on the reshaped outputs is performed to obtain the correlation matrix. Exponential and normalization terms are implemented by taking {\em softmax} to the rows of the correlation matrix.

It is worth noting that this form of $f$ is essentially the same as the {\em Self-Attention Module} in \citep{Vaswani2017Attention} for machine translation. For a given $i$, $\frac{1}{\bm{C}(X)_i} f(\bx_{i},\bx_{j})$ becomes a {\em softmax} computation along the dimension $j$. Eq.~(\ref{equation:nonlocal_roi}) results in the self-attention form $Y =\textit{softmax}(X^{T}W_{\phi}^{T}W_{\psi}X)$ in \cite{Vaswani2017Attention}.

The remaining part in non-local RoI operation $g$ is responsible for extracting useful non-local information from the input feature. Following the bottleneck design of \cite{He2016ResNet}, we first use a 1-by-1 convolution to reduce the channel dimension and then a 3-by-3 convolution to take in the spatial information. To further cut down memory cost, a global 2D average pooling is applied. Finally, the pooled feature blob of shape $(N, D_g, 1, 1)$ is tiled around $H, W$ spatial dimensions and is appended to the end of input blob, as shown in Fig.~\ref{fig:NLB}b. A ReLU activation function \cite{Nair2010ReLU} is used between the two convolution layers.

\section{Experiments}
We use COCO~\cite{LinMBHPRDZ14} 2017 dataset to evaluate NL-RoI. The comparison baseline is Mask-RCNN~\cite{He2017MaskRCNN}, one of the state-of-the-art frameworks for detection and segmentation. The official train/val splits in COCO 2017 are essentially equal to the unofficial minival COCO 2014 train/val splits, which are used by Mask-RCNN. We refer to the latest resluts reported in Facebook Research's GitHub repository, called {\em Detectron}~\cite{Detectron2018}. These results are generally equal to or better than the ones given in the published papers.
The experiments are based on a reimplementation of Detectron in PyTorch (the official Detectron is written in Caffe2). 
%
\begin{table}[tbh]
\caption{Evaluation on Faster R-CNN on COCO2017 validation set.}
\label{table:fasterRCNN}
\vspace{-4mm}
\begin{center}
\small
\begin{tabular}{c c c | c c c | c c c}
\multicolumn{2}{c}{Method} & Training schedule
& ${\textbf{AP}}^{box}$ 
& ${\textbf{AP}}^{box}_{50}$ 
& ${\textbf{AP}}^{box}_{75}$
& ${\textbf{AP}}^{box}_{S}$
& ${\textbf{AP}}^{box}_{M}$
& ${\textbf{AP}}^{box}_{L}$\\
\Xhline{2\arrayrulewidth}
\multirow{4}{*}{R-50}
	& \multirow{2}{*}{Baseline}
    	& 1x & 36.71 & 58.45 & 39.61 & 21.12 & 39.85 & 48.13\\
       && 2x & 37.90 & 59.25 & 41.10 & 21.50 & {\bf 41.10} & 49.91\\
    \cline{2-9}
 	& \multirow{2}{*}{NL-RoI} 
    	& 1x & 37.59 & 60.22 & 40.61 & 22.10 & 40.81 & 48.59\\
       && 2x & {\bf 38.40} & {\bf 60.48} & {\bf 41.45} & {\bf 22.91} & 40.98 & {\bf 50.40}\\
\hline
\multirow{4}{*}{R-101}
	& \multirow{2}{*}{Baseline}
    	& 1x & 39.40 & 61.19 & 43.41 & 22.57 & 42.91 & 51.37\\
       && 2x & 39.78 & 61.29 & 43.28 & 22.88 & 43.33 & {\bf 52.65}\\
    \cline{2-9}
 	& \multirow{2}{*}{NL-RoI} 
    	& 1x & 39.72 & {\bf 62.33} & 43.02 & {\bf 23.67} & 43.40 & 51.54\\
       && 2x & {\bf 40.15} & 62.13 & {\bf 43.47} & 23.20 & {\bf 43.66} & 52.54\\
\hline
\end{tabular}
\end{center}
\end{table}

\paragraph{Faster R-CNN on COCO.}
As shown in Table~\ref{table:fasterRCNN}, NL-RoI can achieve around $0.5\%$ improvement in ${\bf AP}^{box}$ either with R-50 or R-101 backbone network.  Similar improvements are achieved using both short (1x) and longer (2x) training schedules. 
NL-RoI makes the training of Faster R-CNN more effective, since the model trained with 1x schedule can still achieve competitive performance using only half the time of the baseline trained with 2x schedule. 

\begin{table}[tbh]
\caption{Evaluation on Mask R-CNN. Tested on COCO2017 validation set.}
\label{table:maskRCNN}
\vspace{-4mm}
\begin{center}
\small
\begin{tabular}{ccc|ccc|ccc}
\multicolumn{2}{c}{Method} & Training schedule
& ${\textbf{AP}}^{box}$ 
& ${\textbf{AP}}^{box}_{50}$ 
& ${\textbf{AP}}^{box}_{75}$
& ${\textbf{AP}}^{box}_{S}$
& ${\textbf{AP}}^{box}_{M}$
& ${\textbf{AP}}^{box}_{L}$ \\
\Xhline{2\arrayrulewidth}
\multirow{4}{*}{R-50}
	& \multirow{2}{*}{Baseline}
    	& 1x & 37.69 & 59.16 & 40.86 & 21.36 & 40.76 & 49.75\\
       && 2x & 38.61 & 59.84 & 42.10 & 22.20 & 41.50 & 50.77\\
    \cline{2-9}
 	& \multirow{2}{*}{\textbf{NL-RoI}} 
    	& 1x & 38.26 & 60.55 & 41.39 & 22.79 & 41.28 & 49.84\\
       && 2x & {\bf 39.18} & {\bf 61.22 }& {\bf 42.94} & {\bf 23.78} & {\bf 42.29} & {\bf 51.26}\\
\hline
\multirow{4}{*}{R-101}
	& \multirow{2}{*}{Baseline}
    	& 1x & 40.01 & 61.80 & 43.67 & 22.55 & 43.40 & 52.69\\
       && 2x & 40.89 & 61.94 & {\bf 44.78} & 23.50 & 44.21 & 53.89\\
    \cline{2-9}
 	& \multirow{2}{*}{\textbf{NL-RoI}}
    	& 1x & 40.53 & 62.68 & 44.24 & {\bf 23.77} & {\bf 44.31} & 52.40\\
       && 2x & {\bf 40.92} & {\bf 62.76} & 44.37 & 23.40 & 44.23 & {\bf 54.07}\\
\hline\Xhline{2\arrayrulewidth}\hline
\multicolumn{2}{c}{Method} & Training schedule
& ${\textbf{AP}}^{mask}$ 
& ${\textbf{AP}}^{mask}_{50}$ 
& ${\textbf{AP}}^{mask}_{75}$
& ${\textbf{AP}}^{mask}_{S}$
& ${\textbf{AP}}^{mask}_{M}$
& ${\textbf{AP}}^{mask}_{L}$ \\[8pt]
\Xhline{2\arrayrulewidth}
\multirow{4}{*}{R-50}
	& \multirow{2}{*}{Baseline}
    	& 1x & 33.86 & 55.81 & 35.82 & 14.86 & 36.34 & 50.85\\
       && 2x & 34.48 & 56.45 & 36.29 & 15.60 & 37.09 & 52.06\\
    \cline{2-9}
 	& \multirow{2}{*}{\textbf{NL-RoI}} 
    	& 1x & 34.35 & 57.07 & 36.14 & 16.09 & 36.90 & 51.26\\
       && 2x & {\bf 35.26} & {\bf 57.84} & {\bf 37.30} & {\bf 16.50} & {\bf 38.08} & {\bf 52.21}\\
\hline
\multirow{4}{*}{R-101}
	& \multirow{2}{*}{Baseline}
    	& 1x & 35.92 & 58.30 & 37.95 & 15.95 & 38.92 & 53.23\\
       && 2x & 36.39 & 58.47 & 38.68 & 16.64 & 39.15 & 54.00\\
    \cline{2-9}
 	& \multirow{2}{*}{\textbf{NL-RoI}} 
    	& 1x & 36.32 & 59.09 & 38.19 & 16.40 & 39.53 & 53.59\\
       && 2x & {\bf 36.64} & {\bf 59.39} & {\bf 38.68} & {\bf 16.75} & {\bf 39.62} & {\bf 55.09}\\
\hline
\end{tabular}
\end{center}
\end{table}

\paragraph{Mask R-CNN on COCO.}
The improvements of NL-RoI on Mask R-CNN models are similar to those on Faster R-CNN. An increment around $0.5\%$ in performance is obtained on both bounding box ${\textbf{AP}}^{box}$ and mask ${\textbf{AP}}^{mask}$. However, on the combination of deeper backbone (R-101) and longer training schedule (2x), NL-RoI brings only about $0.03\%$ and $0.25\%$ improvements in ${\textbf{AP}}^{box}$ and ${\textbf{AP}}^{mask}$, respectively. This phenomenon suggests that deeper neural networks may have better abilities to encode cross objects relations in high-level features if denser information about individual objects, such as instance masks, is available while training. A supporting evidence for this hypothesis can be found in experimental results of Faster R-CNN in Table~\ref{table:fasterRCNN}. When training Faster R-CNN, only the sparse annotations about the objects, \ie, the bounding boxes, are provided, and the improvements on a deeper backbone achieved by NL-RoI are more significant.

Despite less significant improvement on deeper backbone models, NL-RoI still has better average precision over the baselines on almost every metric. On deeper backbone models, again, we can observe the same behavior of alternating first place between two training schedules in each metric, as previously shown in the results of Faster R-CNN. This behavior only exits in the scores for box APs, but not in mask APs. The box head network is composed of two FC layers, i.e., a two-layer MLP. In contrast, the mask head network consists of four convolution layers and one transposed-convolution layer. This discrepancy, as shown in box APs and mask APs of R-101 NL-RoI models, provides another support to the previous discussion about the cause to the behavior: The overpowered high-level features extracted by a deeper backbone saturate the head network and limit its capacity.

\section{Conclusion}
Non-local RoI is a generic module to improve the performance of R-CNN based methods by explicitly modeling the relations and attention mechanisms between different object regions. Through the experiments on COCO dataset, we show that NL-RoI achieves consistent improvements on Faster R-CNN and Mask R-CNN with different backbone networks and training schedules. Althogh the experimental results also indicate that, when denser or more detailed annotations about objects such as segmentations are given during training, deep neural networks may have the ability to learn object relations implicitly to some extent, we show that using NL-RoI to model the relations between objects in perceptual tasks is still more effective and advantageous.

\bibliographystyle{plainnat}
\bibliography{nlroi}

\section*{Appendix A: Training and Inference}
All the models presented in the paper are end-to-end trained and the residual backbone network is initialized with pretrained weights for ImageNet classification. Batch normalization is not used during training, and the parameters of batch normalization layers, \ie, moving means and moving variances, are merged into only two factors: scaling and shift. All batch normalization layers in the original backbone network are replaced with simple affine transformation layers, which is also done by {\em Detectron}.

The aspect ratios of input images are not change, but the size is rescaled to 800 pixels on the shorter side. If the length of the longer side after rescaling exceeds 1{,}333 pixels, the image is further resized to make sure the length of the longer side is 1{,}333 pixels. For preparing training batches, the images that are to be placed on the same GPU are padded to the maximum height and width of them all. Therefore, image batches to different GPUs may have different padded sizes. We group the images by their aspect ratios so that we can have more compact padded image batches for better occupancy of GPU memory.

NL-RoI is applied to two residual backbones, R-50 and R-101, of different numbers of 50 and 101 convolution layers, respectively. Feature Pyramid Networks~\cite{Lin2017FPN} are used on both cases. As for optimization, stochastic gradient descent with momentum 0.9 and weight decay 0.0001 is used. There are two training schedules available for our experiments. First, the \textbf{1x} schedule starts with a learning rate of 0.02, then reduces it by a factor of 10 at the $60k^{th}$ and the $80k^{th}$ iteration, and has $90k$ iterations in total. Second, the \textbf{2x} schedule also starts with the same learning rate of 0.02, but reduces it by a factor of 10 at the $120k^{th}$ and the $160k^{th}$ iteration, and has $180k$ iterations in total.

A score threshold of 0.05 and greedy non-maximal suppression (NMS) are used to produce the final detection. NMS is only applied among predictions of same category and the suppression threshold is 0.5. A maximum number of 1000 object proposals is used for RPN.

\section*{Appendix B: Ablation Study}
\setcounter{table}{0} \renewcommand{\thetable}{B\arabic{table}}
According to the implementation by \citet{Vaswani2017Attention}, the relation scores computed by two feature vectors are normalized using the square root of feature length. 
In our implementation, the relation scores are computed as the flattened features derived from 3-dimensional tensors. Therefore, either the length of flattened feature or the channel dimension of original tensor could be used for the scaling factor. We also study the effect of applying or not applying the attention mechanism to the same RoI. That is, by setting the diagonal values of the N-by-N relation score matrix to zero, the embedded RoI features from the same RoI will not contribute in the non-local features extracted by NL-RoI. As shown in Table~\ref{table:ablation}, allowing ``attend to self'' and using only channel dimension in scaling factor can achieve the best performance on Faster R-CNN. We choose the last configuration in Table~\ref{table:ablation} as the standard setting for our experiments.

\begin{table}[thb]
\caption{Ablation study on the implementation of NL-RoI based on Faster R-CNN. The configuration in the fourth row is adopted as the standard setting for our experiments.}
\label{table:ablation}
\vspace{-4mm}
\begin{center}
\small
\begin{tabular}{c c|c c c|c c c}
\thead{Attend to\\self} & Scaling Factor 
& ${\textbf{AP}}^{box}$ 
& ${\textbf{AP}}^{box}_{50}$ 
& ${\textbf{AP}}^{box}_{75}$
& ${\textbf{AP}}^{box}_{S}$
& ${\textbf{AP}}^{box}_{M}$
& ${\textbf{AP}}^{box}_{L}$\\
\Xhline{2\arrayrulewidth}
No & $\sqrt{D_f \times H \times W}$ & 36.96 & 59.21 & 39.84 & 20.43 & 40.29 & \bf{49.39} \\\hline
No & $\sqrt{D_f}$ & 37.32 & 59.82 & 39.97 & \bf{22.24} & 40.31 & 48.72 \\\hline
Yes & $\sqrt{D_f \times H \times W}$ & 37.52 & 60.12 & 40.50 & 22.13 & 40.45 & 49.19 \\\hline
Yes & $\sqrt{D_f}$ & \bf{37.59} & \bf{60.22} & \bf{40.61} & 22.10 & \bf{40.81} & 48.59\\\hline
\end{tabular}
\end{center}
\end{table}

\end{document}